\documentclass[11pt]{article}

\usepackage[preprint]{acl}

\usepackage{times}
\usepackage{latexsym}

\usepackage[T1]{fontenc}

\usepackage[utf8]{inputenc}

\usepackage{microtype}

\usepackage{inconsolata}

\usepackage{graphicx}

\title{The Multilingual Quantization Tax: Structural Collapse and Typological Fragility in Edge SLMs}

\author{Mohammad Wathiq Soualhi \\
  Independent Researcher \\ 
  Kuala Lumpur, Malaysia \\
  \texttt{wathiqclaw@gmail.com}}

\usepackage{amsmath}
\usepackage{booktabs}
\usepackage{multirow}
\usepackage{subcaption}
\begin{document}
\maketitle
\begin{abstract}
While 4-bit weight quantization is critical for deploying Small Language Models (SLMs) on edge devices, evaluations of the resulting performance degradation---the quantization tax---remain overwhelmingly English-centric. We present a zero-shot multilingual evaluation of 4-bit quantization across the Gemma 4 and Qwen 3.5 architectures. Evaluating on eight typologically diverse languages using MMLU Pro X Lite and Global PIQA, we show parameter truncation exposes deep pre-training inequalities. We identify four phenomena: (1) \textbf{Typological Fragility:} low-resource and specific non-Latin scripts suffer representational collapse via architecture-specific double dissociations, failing to generate valid task logits; (2) \textbf{Home Language Fragility Paradox:} foundational pre-training pathways provide limited precision loss protection; (3) \textbf{Domain-Specific Forgetting:} multi-step cross-lingual routing degrades while associative soft-science recall remains robust; and (4) \textbf{Quantization Resistance:} highly saturated, typologically aligned domains resist deterministic degradation, with post-quantization performance gains bounded by statistical noise.
\end{abstract}

\section{Introduction}
Deploying edge-based SLMs relies on 4-bit weight quantization \citep{11447893, dettmers2023qloraefficientfinetuningquantized}, yet evaluations of this performance degradation---which we term quantization tax---remain largely English-centric \citep{marchisio-etal-2024-quantization}. Treating this absolute drop in zero-shot accuracy as a monolithic penalty obscures critical vulnerabilities: precision loss uniquely degrades underrepresented languages, non-Latin scripts, and the delicate internal routing mechanisms required for cross-lingual transfer \citep{marchisio-etal-2024-quantization}.

In this paper, we evaluate zero-shot multilingual reasoning in state-of-the-art SLMs under 4-bit NormalFloat (\texttt{nf4}) quantization. Using calibration-free \texttt{nf4} avoids the English-centric data bias inherent in post-training methods like AWQ \citep{MLSYS2024_42a452cb} or GPTQ \citep{frantar2023gptqaccurateposttrainingquantization}, isolating the pure structural impact of parameter truncation \citep{chimoto-etal-2026-calibrating,marchisio-etal-2024-quantization}. We explicitly disable internal thinking modes to prevent models from using test-time compute to bypass quantized bottlenecks, allowing us to evaluate their intrinsic structural integrity. By evaluating the 2B and 4B instruction-tuned variants of Gemma 4 \citep{google_2026} and Qwen 3.5 \citep{yang2025qwen3technicalreport} across eight typologically diverse languages on MMLU Pro X Lite \citep{xuan2025mmluproxmultilingualbenchmarkadvanced} and Global PIQA \citep{chang2025globalpiqaevaluatingphysical}, we map the degradation boundaries between formal logic and associative recall. Our core contributions identify:

\textbf{Typological Fragility:} Quantization tax is highly unequal and exposes a cross-architectural double dissociation. Specific non-Latin scripts (e.g., Hindi, Arabic) experience critical structural collapse depending entirely on the model's distinct pre-training mixture, losing the capacity to generate valid task logits.

\textbf{Home Language Fragility Paradox}: Heavily optimized pathways dedicated to a model's foundational pre-training languages (e.g., English for Gemma, Chinese for Qwen) paradoxically lack structural immunity, degrading at rates comparable to shallower multilingual representations.

\textbf{Domain-Specific Forgetting}: Multi-step logical reasoning is severely impaired. We hypothesize quantization disrupts the outlier-dependent cross-lingual routing required to translate queries into English-centric reasoning cores, while native associative recall remains intact.

\textbf{Quantization Resistance}: Highly saturated, typologically aligned associative domains are structurally resistant to cross-lingual collapse. We show via standard error analysis that apparent post-quantization performance gains represent stochastic variance rather than active regularization.

\section{Experimental Setup}
We designed a zero-shot framework to evaluate the multilingual quantization tax across instruction-tuned variants of Gemma 4 (E2B-it/E4B-it) and Qwen 3.5 (2B/4B). All evaluations were executed via lm-evaluation-harness (v0.4.12) \citep{biderman2024lessonstrenchesreproducibleevaluation} with the \texttt{vLLM} (v0.20.1) \citep{10.1145/3600006.3613165} backend, applying the models' native chat templates.

\textbf{Quantization Strategy} We compare native \texttt{bfloat16} baselines against \texttt{nf4} quantized counterparts via \texttt{bitsandbytes} (v0.45.3) \citep{dettmers2023qloraefficientfinetuningquantized}. Calibration-free \texttt{nf4} was selected to isolate the pure architectural impact of weight truncation, avoiding linguistic biases injected by English-dominant calibration datasets in other quantization methods such as AWQ or GPTQ. To assess raw autoregressive zero-shot priors, internal thinking modes were disabled for all models.

\textbf{Evaluation Domains} To expose pre-training inequalities, we evaluate eight typologically diverse languages (English, Arabic, Russian, Chinese, Japanese, Hindi, Swahili, and Yoruba) across two distinct cognitive benchmarks. The first, \textbf{MMLU Pro X Lite} \citep{xuan2025mmluproxmultilingualbenchmarkadvanced}, is a multidisciplinary reasoning benchmark (10-choice format, 10\% random baseline). Following \citet{xuan2025mmluproxmultilingualbenchmarkadvanced}, we use the Lite version for computational efficiency; it maintains consistent relative model rankings with a negligible performance gap compared to the full corpus. The second, \textbf{Global PIQA} \citep{chang2025globalpiqaevaluatingphysical}, assesses physical commonsense reasoning (binary choice, 50\% baseline) using both \textit{Parallel} (translated) and \textit{Non-parallel} (native, culturally contextualized) subsets to probe internal cross-lingual routing mechanisms. We retained the framework's default task decoding: greedy decoding ($T=0.0$) for MMLU Pro X Lite and nucleus sampling ($T=0.8, p=0.95$) for Global PIQA. 

\begin{table*}[t]
\centering
\small
\begin{tabular}{llcccccccc}
\toprule
\textbf{Model} & \textbf{State} & \textbf{English} & \textbf{Chinese} & \textbf{Russian} & \textbf{Japanese} & \textbf{Arabic} & \textbf{Hindi} & \textbf{Swahili} & \textbf{Yoruba} \\
\midrule
\multirow{3}{*}{\textbf{Gemma 4 E2B-it}} 
& Base (\texttt{bf16}) & 57.3 & 39.6 & 51.4 & 49.2 & 36.9 & 49.2 & 35.9 & 13.3 \\
& Quant (\texttt{nf4}) & 52.4 & 39.5 & 44.7 & 44.2 & 36.2 & 44.9 & 30.4 & 14.3 \\
& \textbf{Tax ($\Delta$)} & \textbf{4.9} & \textbf{0.1} & \textbf{6.7} & \textbf{5.0} & \textbf{0.7} & \textbf{4.3} & \textbf{5.5} & \textbf{-1.0*} \\
\midrule
\multirow{3}{*}{\textbf{Gemma 4 E4B-it}} 
& Base (\texttt{bf16}) & 65.5 & 54.4 & 58.5 & 59.0 & 48.6 & 56.8 & 49.2 & 23.0 \\
& Quant (\texttt{nf4}) & 61.2 & 50.3 & 52.9 & 53.2 & 37.2 & 52.9 & 42.0 & 18.2 \\
& \textbf{Tax ($\Delta$)} & \textbf{4.3} & \textbf{4.1} & \textbf{5.6} & \textbf{5.8} & \textbf{11.4} & \textbf{3.9} & \textbf{7.1} & \textbf{4.8} \\
\midrule
\multirow{3}{*}{\textbf{Qwen 3.5 2B}} 
& Base (\texttt{bf16}) & 50.3 & 44.9 & 23.6 & 34.9 & 34.7 & 1.0 & 11.1 & 16.3 \\
& Quant (\texttt{nf4}) & 44.9 & 36.4 & 7.5 & 27.4 & 27.2 & 0.5 & 1.2 & 5.6 \\
& \textbf{Tax ($\Delta$)} & \textbf{5.4} & \textbf{8.5} & \textbf{16.1} & \textbf{7.5} & \textbf{7.5} & \textbf{0.5*} & \textbf{9.9*} & \textbf{10.7} \\
\midrule
\multirow{3}{*}{\textbf{Qwen 3.5 4B}} 
& Base (\texttt{bf16}) & 66.0 & 61.9 & 61.2 & 59.0 & 55.4 & 3.6 & 29.9 & 3.6 \\
& Quant (\texttt{nf4}) & 66.2 & 58.3 & 62.8 & 55.1 & 52.7 & 1.5 & 23.0 & 2.4 \\
& \textbf{Tax ($\Delta$)} & \textbf{-0.2} & \textbf{3.6} & \textbf{-1.6} & \textbf{3.9} & \textbf{2.7} & \textbf{2.1*} & \textbf{7.0} & \textbf{1.2*} \\
\bottomrule
\end{tabular}
\caption{The Quantization Tax across typologically diverse languages on MMLU Pro X Lite. All values represent percentages (\%). $\Delta$ represents the absolute percentage drop. (*) Models operating within the margin of the 10\% random-chance baseline experience structural logit collapse or stochastic variance.}
\label{tab:hero_tax_comprehensive}
\end{table*}

\section{Results and Analysis}
We frame our analysis around the quantization tax: the absolute drop in zero-shot accuracy ($\Delta = \text{Acc}_{\text{full}} - \text{Acc}_{\text{quantized}}$).

\subsection{Architectural Robustness, Scale, and Resistance}
In \texttt{bfloat16} baselines, Gemma architectures average 46.72\% on MMLU Pro X Lite, while Qwen averages 34.84\% (averages derived from Table~\ref{tab:hero_tax_comprehensive}). Despite this higher baseline, Gemma exhibits superior resilience, incurring an average aggregate quantization tax of only 4.56\% versus Qwen's 5.30\%. Model scale also provides a protective buffer in aggregate; 2B models suffer significantly higher degradation ($\Delta = 5.76\%$) than 4B models ($\Delta = 4.10\%$, $p = 0.036$ via paired t-test across all granular language-domain subtasks, $N=224$). However, as explored in Section 3.2, this protective effect is bounded by baseline script viability.

Certain associative domains exhibit \textit{quantization resistance}; however, apparent performance gains (negative tax) represent statistical noise rather than active regularization. As detailed in Appendix \ref{sec:appendix_global_piqa}, standard error analysis confirms that these fluctuations, whether at the highly competent ceiling (e.g., Gemma 4 E4B-it's $+1.0\%$ on native Hindi PIQA) or the structurally collapsed floor (e.g., Gemma 4 E2B-it's $+1.0\%$ on Yoruba MMLU Pro X Lite), operate within the margin of stochastic variance ($|\Delta| < \sqrt{\text{SE}_{\text{base}}^2 + \text{SE}_{\text{quant}}^2}$). Thus, 4-bit truncation flattens network stability into noise at both performance extremes.

\subsection{Typological Fragility and Structural Collapse}
As detailed in Table \ref{tab:hero_tax_comprehensive}, quantization exposes the typological biases of pre-training corpora. High-resource Latin-script languages (English) exhibit robust penalties ($-0.2\%$ to $5.4\%$). In contrast, low-resource and specific non-Latin scripts face severe performance drops. Scale reduces aggregate degradation (Section 3.1), but cannot rescue languages lacking baseline script viability, acting as a competence multiplier rather than substituting for missing typological alignments. Low-resource and non-Latin scripts frequently suffer critical failure rates, dropping well below the MMLU Pro X Lite 10\% random-chance baselines (e.g., Qwen 3.5 4B failing on Hindi and Yoruba at 3.6\%). While low-resource languages typically suffer from high token fertility \citep{ahia-etal-2023-languages}, sequence truncation is not a factor here: prompts fall comfortably within both architectures' context windows, and generative targets remain single classification tokens. Generation logs confirm that this sub-random performance indicates representational collapse: the quantized model loses the structural fidelity to map highly fertile subword sequences into valid target logits, generating non-target tokens instead.

However, structural vulnerability also degrades highly competent representations, exposing stark architectural discrepancies. While Gemma maintains stability on Cyrillic (Gemma 4 E4B-it incurs a 5.6\% tax on Russian), Qwen's architecture suffers its worst structural degradation on the same script: Qwen 3.5 2B plummets from 23.6\% to 7.5\% ($\Delta = 16.1\%$). This disparity aligns with known differences in tokenizer fertility. Non-Latin scripts like Cyrillic experience subword over-fragmentation \citep{ahia-etal-2023-languages,Petrov-tokenfertility}, compounding precision loss across artificially lengthened token sequences and exacerbating Qwen's structural degradation relative to Gemma.

This non-Latin vulnerability exposes a cross-architectural typological dissociation. Gemma 4 E4B-it incurs an 11.4\% tax on Arabic MMLU Pro X Lite—its largest architectural degradation—while maintaining stability on Hindi ($\Delta = 3.9\%$). Conversely, Qwen 3.5 4B suffers structural collapse on Hindi (failing at a 3.6\% baseline) but resists degradation on Arabic ($\Delta = 2.7\%$). This double dissociation demonstrates that typological fragility is not a uniform penalty for non-Latin scripts, but a direct consequence of each architecture's distinct tokenizer vocabulary allocation and pre-training mixture. Furthermore, this vulnerability is triggered by logical complexity; on associative commonsense tasks (native Global PIQA), Gemma's Arabic performance drops by only 1.0\% (as detailed in Appendix \ref{sec:appendix_global_piqa}). We hypothesize this vulnerability stems from pre-training concentrating critical linguistic routing into high-magnitude outlier weights. As established by \citet{10.5555/3600270.3602468}, emergent outliers govern core reasoning pathways and are disproportionately corrupted by low-bit truncation like \texttt{nf4}. Peripheral languages likely rely on distributed, lower-magnitude representations that fall safely under the quantization clipping threshold.

Paradoxically, deep corpus representation does not grant structural immunity—the Home Language Fragility Paradox. Counterintuitively, foundational pathways resist parameter truncation no better than shallower representations. While Qwen 3.5 4B's bilingual foundation grants its English pathways quantization resistance on MMLU Pro X Lite ($\Delta = -0.2\%$), its primary foundational pillar, Chinese, incurs a highly notable 3.6\% tax (See Table \ref{tab:hero_tax_comprehensive}). This divergence suggests that \texttt{nf4} parameter truncation favors Latin-script token alignments, meaning pre-training volume alone cannot overcome typological incompatibility. The lack of a protective buffer is a cross-architectural phenomenon: the highly English-optimized Gemma 4 E4B-it suffers an English tax ($\Delta = 4.3\%$) that is highly comparable to its degradation on shallower representations like Hindi ($\Delta = 3.9\%$). This suggests highly dense latent pathways remain fundamentally unshielded from precision loss, degrading at rates similar to peripheral languages.

\subsection{Domain Forgetting and Cross-Lingual Routing} 
\begin{figure}[t]
    \centering
    \includegraphics[width=\linewidth]{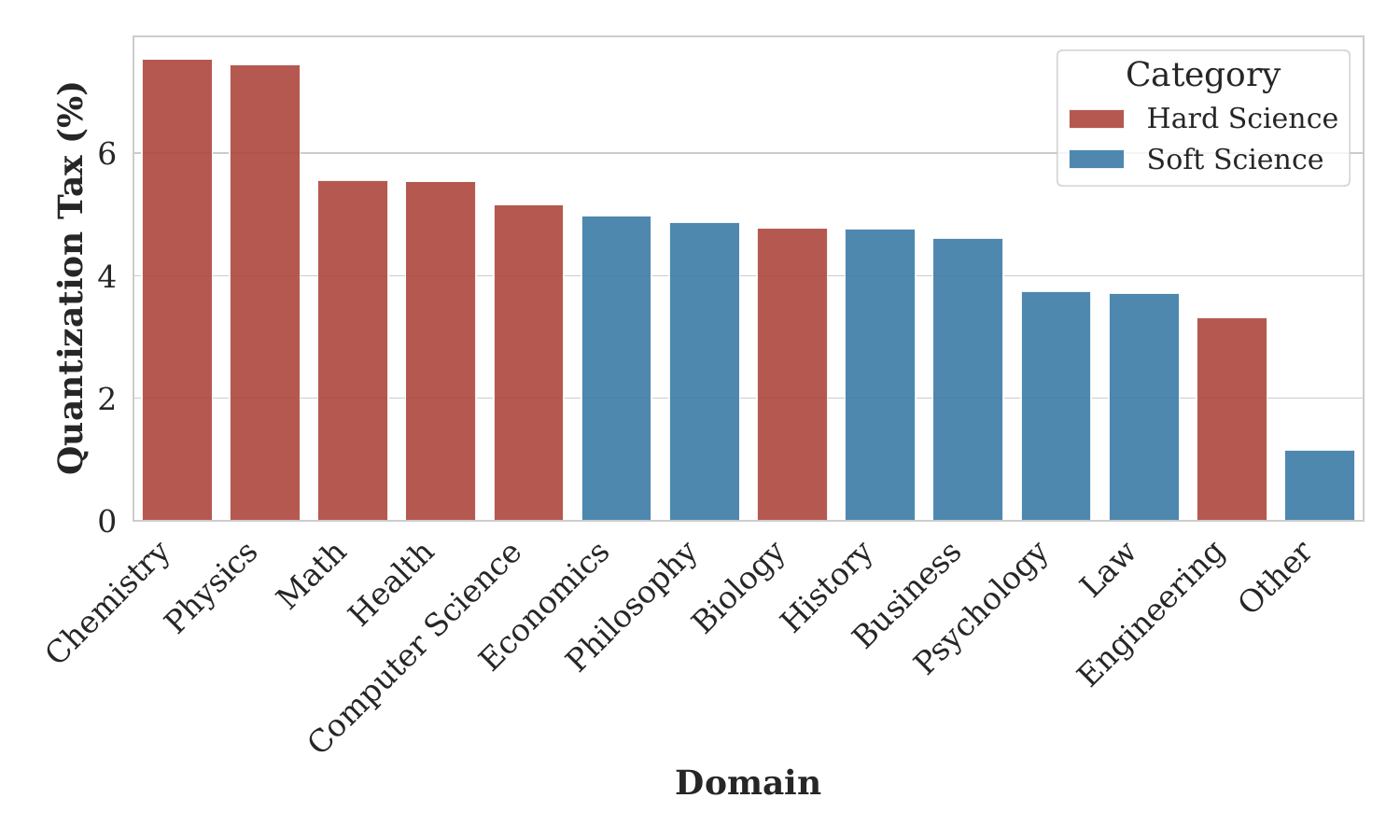}
    \caption{Aggregate domain-specific quantization tax ($\Delta$) on MMLU Pro X Lite, averaged across all models. Hard sciences experience severe degradation, while associative soft sciences remain relatively resilient.}
    \label{fig:domain_forgetting}
\end{figure}

\begin{figure}[t]
    \centering
    \includegraphics[width=\linewidth]{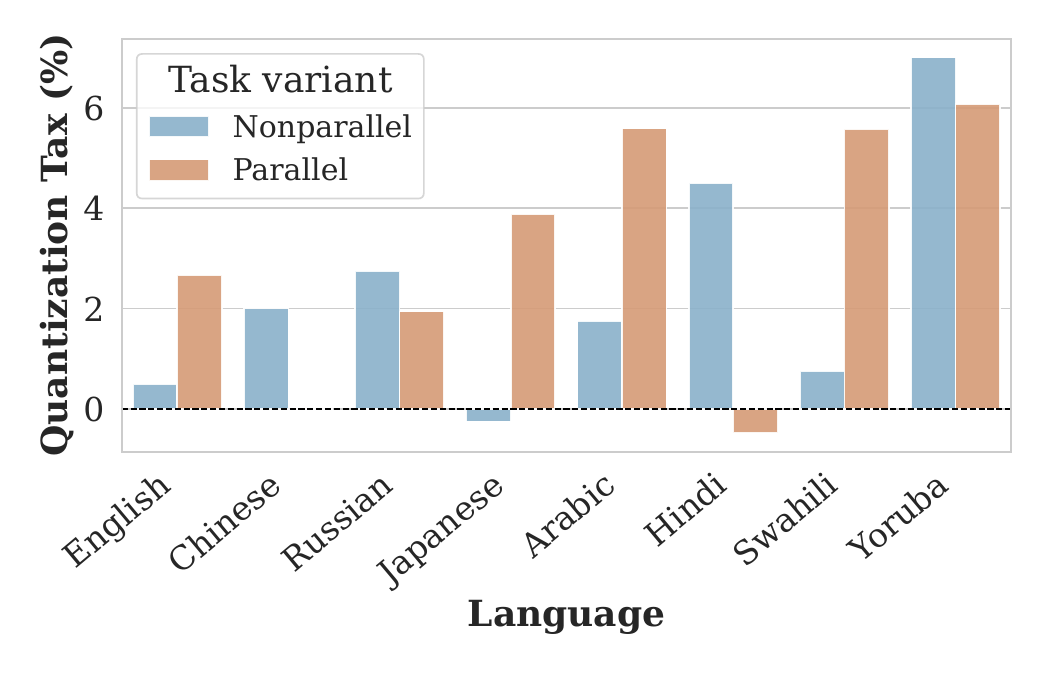}
    \caption{Aggregate Global PIQA quantization tax ($\Delta$) averaged across all models, showing the fragility of Parallel (translated) cross-lingual routing versus stable Nonparallel (native) reasoning.}
    \label{fig:cross_lingual}
\end{figure}

Quantization disproportionately impairs tasks requiring multi-step symbolic manipulation. Figure~\ref{fig:domain_forgetting} illustrates the aggregate domain-specific quantization tax on MMLU Pro X Lite, averaged across all models. Hard sciences suffer greater degradation: Chemistry ($\Delta = 7.53\%$), Physics ($\Delta = 7.45\%$), and Mathematics ($\Delta = 5.56\%$). Conversely, soft sciences reliant on associative recall (e.g., Law, Psychology) incur roughly half the tax ($\Delta \approx 3.7\%$), while others like Economics and Business degrade closer to 4.8\%. Engineering acts as an anomaly ($\Delta = 3.3\%$) among the sciences; it heavily tests factual recall rather than multi-step logic, leveraging quantization-resistant associative pathways. 
This disparity between logical and associative resilience is corroborated by the fragility of internal cross-lingual alignment on Global PIQA. As shown in Figure \ref{fig:cross_lingual}, aggregate performance diverges sharply between translated (Parallel) and native (Nonparallel) reasoning pathways. Note that Parallel Hindi's apparent aggregate resilience is an artifact of stochastic variance; Qwen baselines operate near or below the 50\% random-chance floor on this task, reducing fluctuations to meaningless noise.  However, quantization resistance is resource- and scale-dependent. While high-resource native reasoning on Global PIQA (e.g., Japanese non-parallel) incurs negligible degradation, models require a minimum threshold of pre-training saturation to maintain immunity. Low-resource native pathways degrade, evidenced by Qwen 3.5 4B's native Yoruba Tax ($\Delta = 8.0\%$). At the 2B scale, this structural failure accelerates: Qwen 3.5 2B's native Yoruba falls from 54\% to 40\%  ($\Delta = 14\%$). This absolute drop exceeds its translated Yoruba tax ($\Delta = 8.74\%$, falling from 30.10\% to 21.36\%); however, the translated baseline already operates below the 50\% random-chance floor, indicating pre-existing logit collapse. Therefore, below critical parameter thresholds, even translation-free, culturally aligned associative pathways lose structural integrity under 4-bit quantization.
For mid-to-low resource languages, quantization acts as an acute disruption against the translation layer. Swahili provides a stark example: On Global PIQA, Qwen 3.5 4B's native (Non-parallel) Swahili reasoning drops by a negligible 1.00\% post-quantization. However, on the translated (Parallel) subset, it suffers a massive 11.65\% tax. This extreme divergence strongly implies that quantization does not erase underlying multilingual semantic knowledge; instead, it severs the delicate cross-lingual routing mechanisms required to translate queries into the model's English-centric reasoning core. We hypothesize this routing relies on high-magnitude outlier weights to bridge distinct semantic spaces, which are severely clipped by \texttt{nf4} truncation.

\section{Conclusion}
Deploying SLMs on edge devices increasingly relies on 4-bit quantization, but our evaluation of the Gemma 4 and Qwen 3.5 architectures reveals its impact is far from uniform. The quantization tax exposes inherent pre-training inequalities. We demonstrate that while saturated, high-resource associative pathways exhibit deep Quantization Resistance, the degradation is deeply asymmetric. Through Typological Fragility and Domain-Specific Forgetting, low-resource non-Latin scripts and multi-step cross-lingual routing face severe structural collapse. Furthermore, the Home Language Fragility Paradox suggests that heavily optimized native representations are paradoxically more vulnerable to precision loss. Moving forward, the NLP community must abandon aggregate, English-centric degradation metrics. Equitable global deployment of SLMs demands typologically calibrated and domain-aware quantization strategies.

\section*{Limitations}
While this study exposes critical vulnerabilities in the multilingual capabilities of quantized SLMs, several boundaries to our experimental design must be acknowledged. 

First, our evaluation is bounded by the architectural scope of the Gemma 4 and Qwen 3.5 families. While these models represent the current state-of-the-art in edge-deployable SLMs, architectures from other model families (e.g., Llama, Phi) with different attention mechanisms, tokenizer vocabularies, or foundational pre-training mixtures may exhibit differing profiles of quantization resistance. Furthermore, our analysis is restricted to the 2B and 4B parameter scales; the scaling dynamics of structural collapse in larger edge models (e.g., 7B-8B) remain unexplored.

Second, to isolate the pure architectural impact of parameter truncation, we exclusively utilized calibration-free \texttt{nf4} quantization. Consequently, this study does not evaluate Post-Training Quantization (PTQ) methods requiring calibration datasets, such as AWQ or GPTQ. While current PTQ methods frequently rely on English-centric data—which would obscure the native structural degradation we sought to measure—the application of explicitly culturally-balanced, multilingual calibration datasets might mitigate some of the cross-lingual routing collapse observed in our results. 

Third, our methodology enforces a zero-shot, autoregressive setting with internal thinking modes disabled. We selected this constraint to measure the models' intrinsic structural integrity without the interference of test-time compute. However, real-world edge deployments can often leverage few-shot prompting, system instructions, or reasoning tokens. It is entirely possible that granting models test-time compute could allow them to dynamically bypass corrupted semantic pathways and recover a portion of the incurred quantization tax. 

Finally, our evaluation is restricted to structured reasoning and commonsense classification benchmarks. Open-ended generative tasks, such as open-ended dialogue, summarization, or long-form machine translation, were excluded. The impact of typological fragility on generative fluency—specifically regarding token repetition, hallucination rates, and semantic drift post-quantization—requires dedicated future investigation.
\bibliography{custom}

\appendix
\section{Standard Error Analysis of Quantization Resistance}
\label{sec:appendix_a}

This appendix provides the detailed standard error analysis referenced in Section 3.1, confirming that instances of apparent performance improvement post-quantization are not architectural regularizations, but fall within the bounds of expected stochastic variance ($|\Delta| < \sqrt{\text{SE}_{\text{base}}^2 + \text{SE}_{\text{quant}}^2}$).

\begin{table*}[!htbp] 
\centering
\resizebox{\textwidth}{!}{%
\begin{tabular}{llccccc}
\toprule
\textbf{Model} & \textbf{Task (Dataset \& Lang.)} & \textbf{Base Acc ($\pm$ SE)} & \textbf{Quant Acc ($\pm$ SE)} & \textbf{Tax ($\Delta$)} & \textbf{Bound} & \textbf{Strictly Bounded?} \\
\midrule
\textbf{Gemma 4 E4B-it} & Global PIQA (Native Hindi) & 92.00\% ($\pm$ 2.73\%) & 93.00\% ($\pm$ 2.56\%) & -1.00\% & 3.74\% & \textbf{Yes} \\
\textbf{Gemma 4 E2B-it} & MMLU Pro X Lite (Yoruba) & 13.27\% ($\pm$ 1.38\%) & 14.29\% ($\pm$ 1.44\%) & -1.02\% & 1.99\% & \textbf{Yes} \\
\textbf{Qwen 3.5 4B} & MMLU Pro X Lite (Russian) & 61.22\% ($\pm$ 1.90\%) & 62.76\% ($\pm$ 1.86\%) & -1.54\% & 2.66\% & \textbf{Yes} \\
\textbf{Qwen 3.5 4B} & MMLU Pro X Lite (English) & 65.99\% ($\pm$ 1.87\%) & 66.16\% ($\pm$ 1.87\%) & -0.17\% & 2.64\% & \textbf{Yes} \\
\textbf{Gemma 4 E2B-it} & MMLU Pro X Lite (Chinese) & 39.63\% ($\pm$ 1.93\%) & 39.46\% ($\pm$ 1.92\%) & +0.17\% & 2.72\% & \textbf{Yes} \\
\bottomrule
\end{tabular}}
\caption{Stochastic Variance Bounding for Noise-Level Fluctuations (Including Negative Tax). $|\Delta|$ represents the absolute change in accuracy. Bound is calculated as the standard error of the difference: $\sqrt{\text{SE}_{\text{base}}^2 + \text{SE}_{\text{quant}}^2}$.}
\label{tab:stochastic_variance_bounding}
\end{table*}

\subsection{Detailed Global PIQA Multilingual Degradation}
\label{sec:appendix_global_piqa}

To support the visual trends observed in Figure~\ref{fig:cross_lingual} and ensure full empirical transparency, Table~\ref{tab:global_piqa_detailed} provides the complete raw accuracy metrics for both Native (Non-parallel) and Translated (Parallel) subsets of the Global PIQA benchmark. 

This tabular data corroborates the cross-lingual routing collapse discussed in Section 3.3. It highlights the severe and divergent quantization tax observed against the translation layer in mid-to-low resource languages, substantiating the specific phenomenon where models like Qwen 3.5 4B retain stable native reasoning (e.g., a negligible 1.00\% tax on Non-parallel Swahili) while suffering massive structural degradation on translated queries (an 11.65\% tax on Parallel Swahili).

\begin{table*}[h]
\centering
\small
\begin{tabular}{llcccccc}
\toprule
\multirow{2}{*}{\textbf{Model}} & \multirow{2}{*}{\textbf{Language}} & \multicolumn{3}{c}{\textbf{Native (Non-parallel) PIQA}} & \multicolumn{3}{c}{\textbf{Translated (Parallel) PIQA}} \\
\cmidrule(lr){3-5} \cmidrule(lr){6-8}
& & \textbf{Base} & \textbf{Quant} & \textbf{Tax ($\Delta$)} & \textbf{Base} & \textbf{Quant} & \textbf{Tax ($\Delta$)} \\
\midrule

\multirow{8}{*}{\textbf{Qwen 3.5 4B}} 
& English & 87.00 & 90.00 & -3.00 & 84.47 & 77.67 & 6.80 \\
& Chinese & 83.00 & 81.00 & 2.00 & 71.84 & 72.82 & -0.98 \\
& Russian & 85.00 & 86.00 & -1.00 & 67.96 & 74.76 & -6.80 \\
& Japanese & 88.00 & 85.00 & 3.00 & 71.84 & 68.93 & 2.91 \\
& Arabic & 67.00 & 69.00 & -2.00 & 65.05 & 66.02 & -0.97 \\
& Hindi & 85.00 & 76.00 & 9.00 & 54.37 & 57.28 & -2.91 \\
& \textbf{Swahili} & \textbf{74.00} & \textbf{73.00} & \textbf{1.00} & \textbf{42.72} & \textbf{31.07} & \textbf{11.65} \\
& Yoruba & 64.00 & 56.00 & 8.00 & 27.18 & 23.30 & 3.88 \\
\midrule

\multirow{8}{*}{\textbf{Gemma 4 E4B-it}} 
& English & 90.00 & 87.00 & 3.00 & 82.52 & 79.61 & 2.91 \\
& Chinese & 79.00 & 76.00 & 3.00 & 81.55 & 80.58 & 0.97 \\
& Russian & 92.00 & 92.00 & 0.00 & 77.67 & 73.79 & 3.88 \\
& Japanese & 92.00 & 93.00 & -1.00 & 81.55 & 78.64 & 2.91 \\
& Arabic & 76.00 & 75.00 & 1.00 & 83.50 & 79.61 & 3.89 \\
& Hindi & 92.00 & 93.00 & -1.00 & 81.55 & 76.70 & 4.85 \\
& Swahili & 88.00 & 87.00 & 1.00 & 69.90 & 68.93 & 0.97 \\
& Yoruba & 58.00 & 48.00 & 10.00 & 31.07 & 25.24 & 5.83 \\
\midrule

\multirow{8}{*}{\textbf{Qwen 3.5 2B}} 
& English & 75.00 & 69.00 & 6.00 & 55.34 & 49.51 & 5.83 \\
& Chinese & 67.00 & 65.00 & 2.00 & 50.49 & 49.51 & 0.98 \\
& Russian & 76.00 & 70.00 & 6.00 & 53.40 & 52.43 & 0.97 \\
& Japanese & 74.00 & 77.00 & -3.00 & 39.81 & 41.75 & -1.94 \\
& Arabic & 60.00 & 48.00 & 12.00 & 52.43 & 39.81 & 12.62 \\
& Hindi & 65.00 & 57.00 & 8.00 & 24.27 & 30.10 & -5.83 \\
& Swahili & 50.00 & 50.00 & 0.00 & 24.27 & 20.39 & 3.88 \\
& Yoruba & 54.00 & 40.00 & 14.00 & 30.10 & 21.36 & 8.74 \\
\midrule

\multirow{8}{*}{\textbf{Gemma 4 E2B-it}} 
& English & 77.00 & 81.00 & -4.00 & 69.90 & 74.76 & -4.86 \\
& Chinese & 72.00 & 71.00 & 1.00 & 65.05 & 66.02 & -0.97 \\
& Russian & 85.00 & 79.00 & 6.00 & 73.79 & 64.08 & 9.71 \\
& Japanese & 83.00 & 83.00 & 0.00 & 75.73 & 64.08 & 11.65 \\
& Arabic & 69.00 & 73.00 & -4.00 & 73.79 & 66.99 & 6.80 \\
& Hindi & 89.00 & 87.00 & 2.00 & 66.99 & 65.05 & 1.94 \\
& Swahili & 85.00 & 84.00 & 1.00 & 63.11 & 57.28 & 5.83 \\
& Yoruba & 53.00 & 57.00 & -4.00 & 30.10 & 24.27 & 5.83 \\
\bottomrule
\end{tabular}
\caption{Raw Quantization Tax ($\Delta$) breakdown between Native (culturally aligned) and Translated (forced multi-step routing) evaluation pathways on the Global PIQA benchmark. All values represent percentages (\%). The Qwen 3.5 4B Swahili discrepancy referenced in the main text is bolded for clarity.}
\label{tab:global_piqa_detailed}
\end{table*}
\clearpage

\section{Computational Environment and Reproducibility}
\label{sec:appendix_reproducibility}
\subsection{Infrastructure and Compute Budget}
All evaluations were executed using the \texttt{lm-evaluation-harness} (v0.4.12) framework. Model inference was accelerated using the \texttt{vLLM} (v0.20.1) backend, and calibration-free 4-bit NormalFloat (\texttt{nf4}) quantization was implemented via \texttt{bitsandbytes} (v0.45.3). The experimental pipeline was deployed on a single NVIDIA RTX A6000 GPU, provisioned via Thunder Compute. In total, the complete suite of multilingual and multi-domain zero-shot evaluations required approximately 14 GPU hours to complete.
\subsection{Evaluation Command Templates}
To ensure full reproducibility of the zero-shot pipeline, evaluations were executed using the following generalized command structure:
\begin{verbatim}
lm_eval --model vllm \
  --model_args pretrained=[MODEL_ID],
  enable_thinking=False,
  trust_remote_code=True,
  tensor_parallel_size=1,
  gpu_memory_utilization=0.9,
  [SPECIFIC_ARGS] \
  --tasks [TASK_LIST] \
  --apply_chat_template \
  --num_fewshot 0 \
  --batch_size auto \
  --output_path ./results_[MODEL_ID] \
  --log_samples
\end{verbatim}
The \texttt{[MODEL\_ID]} parameter corresponds to the specific model identifiers evaluated in this study: \texttt{gemma-4-E2B-it}, \texttt{gemma-4-E4B-it}, \texttt{Qwen3.5-2b}, and \texttt{Qwen3.5-4B}.

The \texttt{[SPECIFIC\_ARGS]} parameter was strictly modulated to enforce architectural constraints and quantization states:

\noindent \textbf{Gemma 4 Baselines:} Left blank. Note that the \texttt{reasoning\_parser} argument was deliberately omitted for Gemma 4, as passing it within this framework version paradoxically re-enables thinking mode, violating our zero-shot autoregressive constraints.\\

\noindent \textbf{Qwen 3.5 Baselines:} \texttt{enable\_thinking=False, reasoning\_parser=qwen3}\\

\noindent \textbf{Quantized Variants:} The respective base arguments were appended with \texttt{quantization=bitsandbytes, load\_format=bitsandbytes}.\\

To guarantee exact evaluation alignment, the \texttt{[TASK\_LIST]} parameter was populated with the following precise comma-separated strings for each benchmark:\\

\noindent \textbf{MMLU Pro X Lite:}
{\footnotesize
\begin{verbatim}
mmlu_prox_lite_en,
mmlu_prox_lite_ar,
mmlu_prox_lite_ru,
mmlu_prox_lite_zh,
mmlu_prox_lite_ja,
mmlu_prox_lite_hi,
mmlu_prox_lite_sw,
mmlu_prox_lite_yo
\end{verbatim}
}

\noindent \textbf{Global PIQA:}
{\footnotesize
\begin{verbatim}
global_piqa_parallel_generation_eng_latn,
global_piqa_nonparallel_generation_eng_latn,
global_piqa_parallel_generation_cmn_hans,
global_piqa_nonparallel_generation_cmn_hans,
global_piqa_parallel_generation_jpn_jpan,
global_piqa_nonparallel_generation_jpn_jpan,
global_piqa_parallel_generation_rus_cyrl,
global_piqa_nonparallel_generation_rus_cyrl,
global_piqa_parallel_generation_arb_arab,
global_piqa_nonparallel_generation_arb_arab,
global_piqa_parallel_generation_hin_deva,
global_piqa_nonparallel_generation_hin_deva,
global_piqa_parallel_generation_yor_latn,
global_piqa_nonparallel_generation_yor_latn,
global_piqa_parallel_generation_swh_latn,
global_piqa_nonparallel_generation_swh_latn
\end{verbatim}
}
\end{document}